\documentclass[runningheads]{llncs}

\usepackage[mobile]{eccv}

\usepackage{eccvabbrv}

\usepackage{graphicx}
\usepackage{booktabs}
\usepackage{microtype}

\usepackage[accsupp]{axessibility}

\usepackage{makecell}
\usepackage{wrapfig}

\usepackage[pagebackref,breaklinks,colorlinks=true,linkcolor=blue,citecolor=blue,urlcolor=blue]{hyperref}

\usepackage{orcidlink}

\begin{document}

\title{MegaScenes: Scene-Level View Synthesis at Scale} 

\titlerunning{MegaScenes}

\makeatletter
\newcommand{\printfnsymbol}[1]{%
  \textsuperscript{\@fnsymbol{#1}}%
}

\author{Joseph Tung\thanks{Equal contribution.} \inst{1} \and
Gene Chou\printfnsymbol{1} \inst{1} \and
Ruojin Cai\inst{1} \and
Guandao Yang\inst{2} \and
Kai Zhang\inst{3} \and
Gordon Wetzstein\inst{2} \and
Bharath Hariharan\inst{1} \and
Noah Snavely\inst{1} 
}

\authorrunning{J.~Tung, G.~Chou et al.}

\institute{Cornell University \and
Stanford University  \and
Adobe Research
}

\maketitle
\vspace{-10pt}
\begin{figure}[h]
    \centering
    \includegraphics[width=\textwidth,trim={0 30pt 0 0},clip]{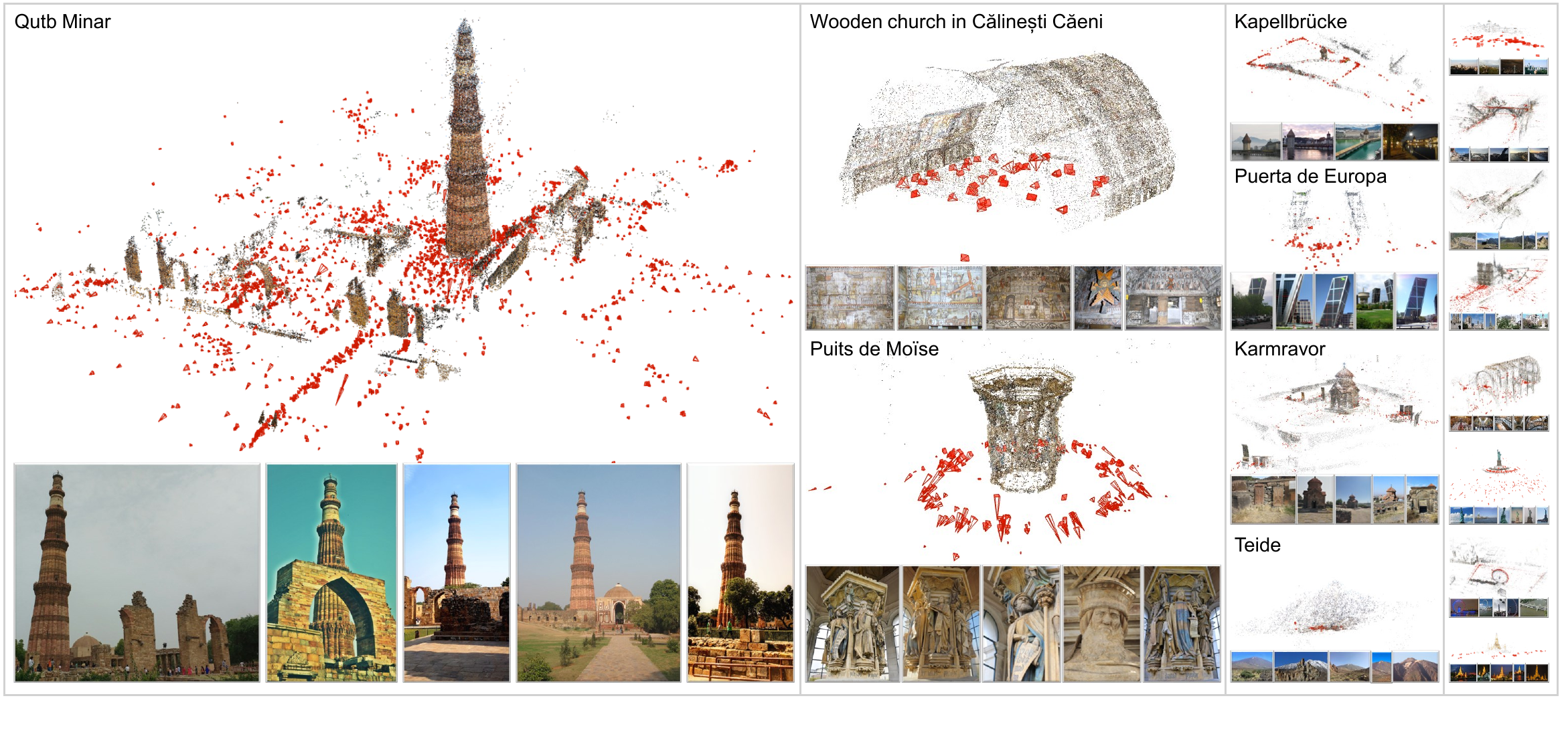}
    \caption{The MegaScenes Dataset is an extensive collection of around 430k scenes, featuring over 100k structure-from-motion reconstructions and over 2 million registered images.
    MegaScenes includes a diverse array of scenes, such as 
    minarets (e.g., Qutb Minar), building interiors (e.g., wooden church in Călinești Căeni), statues (e.g., Puits de Moïse), bridges (e.g., Kapellbrücke), towers (e.g., Puerta de Europa), religious buildings (e.g., Karmravor), and natural landscapes (e.g., Teide volcano).
    The images of these scenes are captured under varying conditions, including different times of day, various weather and illumination, and from different devices with distinct camera intrinsics.}
    \label{fig:teaser}
\end{figure}

\vspace{-20pt}
\begin{abstract}
Scene-level novel view synthesis (NVS) is fundamental to many vision and graphics applications. Recently, pose-conditioned diffusion models have led to significant progress by extracting 3D information from 2D foundation models, but these methods are limited by the lack of scene-level training data. Common dataset choices either consist of isolated objects (Objaverse), or of object-centric scenes with limited pose distributions (DTU, CO3D). 
In this paper, we create a large-scale scene-level dataset from Internet photo collections, called MegaScenes, which contains over 100K structure from motion (SfM) reconstructions from around the world. Internet photos represent a scalable data source but come with challenges such as lighting and transient objects. We address these issues to further create a subset suitable for the task of NVS. Additionally, we analyze failure cases of state-of-the-art NVS methods and significantly improve generation consistency. Through extensive experiments, we validate the effectiveness of both our dataset and method on generating in-the-wild scenes. For details on the dataset and code, see our project page at \url{https://megascenes.github.io}. 
\end{abstract}

\section{Introduction}
Our vast visual experience enables us to look at a single view of a scene and infer what we cannot see. We can see a bridge from afar and imagine what it would be like to stand under it, or view the front of a church and guess what it looks like from other sides.
Imagine a computer vision model that has similarly seen countless scenes: like humans, it can infer other views of a scene from a single image (i.e., it can perform \emph{single-view novel-view synthesis}).
Beyond the connections with human vision, such a vision model would allow us to explore new AR/VR visualizations\cite{RealEstate10k} or plan effectively in robotics\cite{zhou2016view, du2021nerflow}.

Current state-of-the-art on single-view novel-view synthesis takes 2D diffusion models trained on large internet datasets\cite{rombach2022high} and finetunes them on multiview images with camera poses.
Concretely, these finetuned diffusion models map a reference image and a target pose to the corresponding target view~\cite{watson2022novel, liu2023zero}.
These methods successfully produce consistent novel views 
but only at the object level as they were trained on object meshes.
Unfortunately, attempts to generalize this approach to scenes~\cite{chan2023generative, sargent2023zeronvs} by training on existing scene-level datasets~\cite{reizenstein21co3d, RealEstate10k, dtu, infinite_nature_2020} have been held back by the relatively small size and lack of diversity of these scene-level datasets.
As such, current scene-level novel view synthesis techniques cannot match the consistency of object-level models and fail to generalize to realistic, in-the-wild scenes.

To address the lack of diverse, scene-level data for training 3D-aware models, we create MegaScenes, a large-scale 3D dataset. MegaScenes builds on eight million free-to-use images sourced from Wikimedia Commons. We leverage structure from motion (SfM)
to extract 3D structure from internet images at scale. In total, MegaScenes contains over 100K scene-level SfM reconstructions from around the world, along with associated data like captions, as well as the estimated relative poses of tens of millions of image pairs.  \cref{fig:teaser} shows a few example scenes.

While we foresee a variety of 3D-related applications that could benefit from MegaScenes, such as pose estimation~\cite{wang2023pd}, feature matching~\cite{wang2020learning}, and reconstruction~\cite{wang2023dust3r}, in this paper we focus on NVS as a representative application. 
Following prior work in NVS~\cite{liu2023zero, sargent2023zeronvs}, our goal is to generate a plausible image at a target pose given only one reference image.
Therefore, from the MegaScenes dataset we sample image pairs that have consistent lighting and visual overlap to create over 2 million training pairs. 
We validate the effectiveness of MegaScenes by finetuning current state-of-the-art NVS models on our dataset, and find that the new models perform significantly better on multiple dataset benchmarks.

In these experiments, we also identify and mitigate
failure cases of existing methods by including additional conditioning that warps the input image into the target view~\cite{infinite_nature_2020}.
While our method is simple and builds on existing approaches, it addresses the fundamental issues in prior works, and we validate that it produces significantly more consistent and realistic results. 

We show extensive experiments in \cref{sec:nvs} and a large collection of uncurated results in the supplement to demonstrate that our method and our training dataset yield NVS models that are effective across multiple benchmarks.
We will release the dataset, code, and pretrained models. 
\vspace{-1em}

\section{Related Work}

\vspace{-0.5em}
\noindent \textbf{Datasets for 3D Learning.}
Datasets are the keystone of 3D learning.
Recently, many 3D datasets have provided 
increasing amounts of data for tasks such as novel view synthesis, scene understanding, and 3D generation. 
Object-level 3D datasets like ShapeNet~\cite{shapenet}, CO3Dv2~\cite{reizenstein21co3d}, and DTU~\cite{dtu} have been extensively utilized in sparse view NVS~\cite{yu2021pixelnerf} and 3D generation~\cite{zhou2023sparsefusion}. 
The emergence of larger-scale 3D object datasets like MVImgNet~\cite{yu2023mvimgnet} and Objaverse-XL~\cite{objaverseXL} has enabled more generalizable models~\cite{hong2023lrm,wang2023pf,liu2023zero} for 3D reconstruction and generation. However, these datasets are confined to objects and do not extend to full scenes.

At the scene scale, existing datasets~\cite{li2023matrixcity,yao2020blendedmvs, roberts2021hypersim, li2021openrooms, infinite_nature_2020, RealEstate10k, chang2017matterport3d, dai2017scannet, yeshwanth2023scannet++} have facilitated scene-level view synthesis and generation, but are often limited to a constrained set of categories, such as indoor scenes and drone shots of nature.
DL3DV-10K~\cite{ling2023dl3dv} is concurrent work that aims to create a diverse and large-scale 3D scene dataset from videos, but features limited variation in camera poses.

In contrast, scene-level 3D datasets sourced from internet photos, such as MegaDepth~\cite{MegaDepthLi18}, present a diverse distribution of camera poses and intrinsics, various lighting conditions and weather, different times of day, and transient objects and is widely applied in monocular depth estimation~\cite{depthanything, zoedepth} and learned feature matching~\cite{tyszkiewicz2020disk, edstedt2024dedode, lindenberger2023lightglue}. However, MegaDepth
is limited in scale to just
196 landmarks.
Two more recent scene-level datasets include Google Landmarks v2~\cite{weyand2020GLDv2} and WikiScenes~\cite{Wu2021Towers}, which also gather images from Wikimedia Commons. However, Google Landmarks only focuses on 2D retrieval (no 3D information), and WikiScenes focuses on specific categories like cathedrals.

To address these limitations, MegaScenes incorporates diverse scene categories that include indoor, outdoor, natural scenes, and object-like scenes such as statues. 
It significantly extends the scale of 3D scene data, surpassing MegaDepth by several orders of magnitude,
and includes 3D annotations of camera poses and reconstructions.
Sourced from the Wikimedia Foundation, MegaScenes benefits from rich metadata and a wide distribution of illumination and camera poses.
Our findings in novel view synthesis demonstrate that the image diversity within the \textit{same} scene enhances model generalization capabilities, highlighting the value of MegaScenes in advancing the field of 3D learning.

\medskip 
\noindent\textbf{Novel View Synthesis from Sparse Views.}
Novel view synthesis is the task of generating images from unseen views given some known images of a scene.
When many input views are available, one can reconstruct an explicit 3D scene model, e.g., a neural radiance field~\cite{mildenhall2021nerf} or 3D Gaussians~\cite{kerbl20233d}.
However, given only sparse views (or just one), methods must rely on heuristic priors such as geometry smoothness~\cite{verbin2022ref,niemeyer2022regnerf} or data priors~\cite{wu2023reconfusion,yu2021pixelnerf,chan2023generative,tewari2024diffusion}. Recently, a popular line of work uses foundation generative models~\cite{rombach2022high, saharia2022photorealistic} as prior knowledge. To workaround the lack of 3D data and instead use 2D foundation models, \cite{wu2023reconfusion, poole2022dreamfusion, wang2023score,wang2024prolificdreamer} generate 3D objects by enforcing the rendered images from unseen viewpoints to agree with generative models. \cite{chung2023luciddreamer, cai2022diffdreamer, yu2023wonderjourney} explicitly extract multiview images by warping reference images and their depth given target poses following Liu \etal~\cite{infinite_nature_2020}, and use an inpainting model to fill in the missing regions.  
However, since these 2D generative models are not 3D-aware, methods can suffer from artifacts such as the multiface problem~\cite{poole2022dreamfusion} or inconsistent geometries~\cite{chung2023luciddreamer, yu2023wonderjourney}. 

A promising alternative is to leverage generative models that can perform novel-view synthesis conditioned on input view and change of camera poses~\cite{liu2023zero,shi2023mvdream,chan2023generative,watson2022novel,sargent2023zeronvs,hong2023lrm,zhou2023sparsefusion, cai2022diffdreamer, invs2023, liu2024one, liu2023syncdreamer,qian2023magic123}. These methods produce consistent geometry given sparse or even single views without the artifacts from 2D models, and can generalize to unseen scenarios thanks to their data priors. However, these works are generally trained on data with limited diversity, such as on object meshes~\cite{objaverseXL} and object-centric scenes~\cite{reizenstein21co3d}. As we will show later, this limits their applicability to realistic, in-the-wild scenes.
In this paper, we create both a dataset and a method that directly addresses scene-level novel view synthesis. 

\vspace{-1em}
\section{MegaScenes Dataset}
\vspace{-0.5em}
In this section, we introduce the MegaScenes dataset, designed to capture a diverse range of geometries for large-scale scenes---plazas, buildings, interiors, and natural landmarks---using worldwide internet photos.
First, we describe the dataset's key characteristic features in \cref{sec:dataformat}. 
Then, the data collection and reconstruction pipeline of MegaScenes are detailed in \cref{sec:datacurate}
Finally, we provide dataset statistics in \cref{sec:datastats}.

\begin{figure}[t]
    \centering
    \includegraphics[width=\textwidth,trim={0 0 0 0},clip]{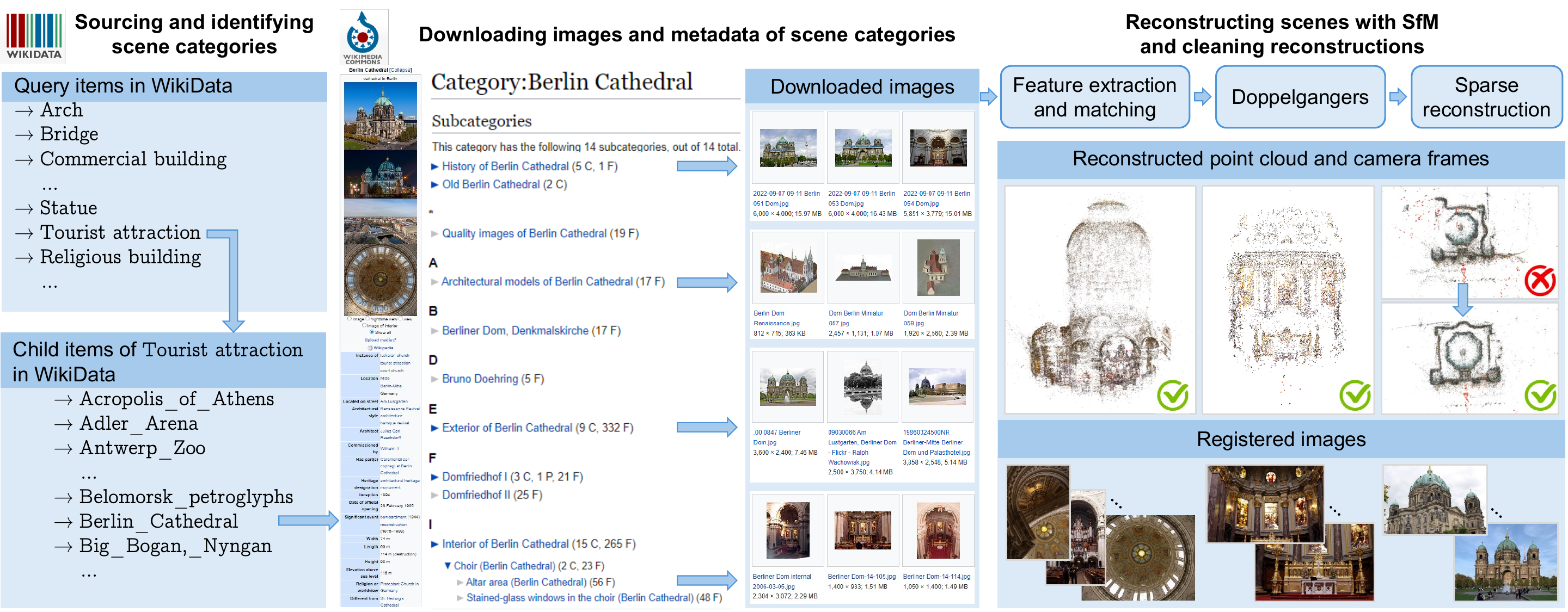}
    \caption{
    MegaScenes curation pipeline. 
    We first source and identify potential scene categories from Wikidata. 
    Subsequently, images and metadata for each scene category is downloaded.
    Finally, we reconstruct scenes using Structure from Motion (SfM) and clean them using the Doppelgangers~\cite{cai2023doppelgangers} pipeline.
    }
    \label{fig:dataset_pipeline}
    \vspace{-1em}
\end{figure}

\vspace{-1em}
\subsection{Dataset Characteristics}
\label{sec:dataformat}

We describe several characteristic features that highlight the versatility of the MegaScenes dataset in training future vision tasks, including category, image, and 3D information.

\begin{figure}[t]

    \centering
    \includegraphics[width=\textwidth, trim={0 0.98cm 0.5cm 0}, clip]{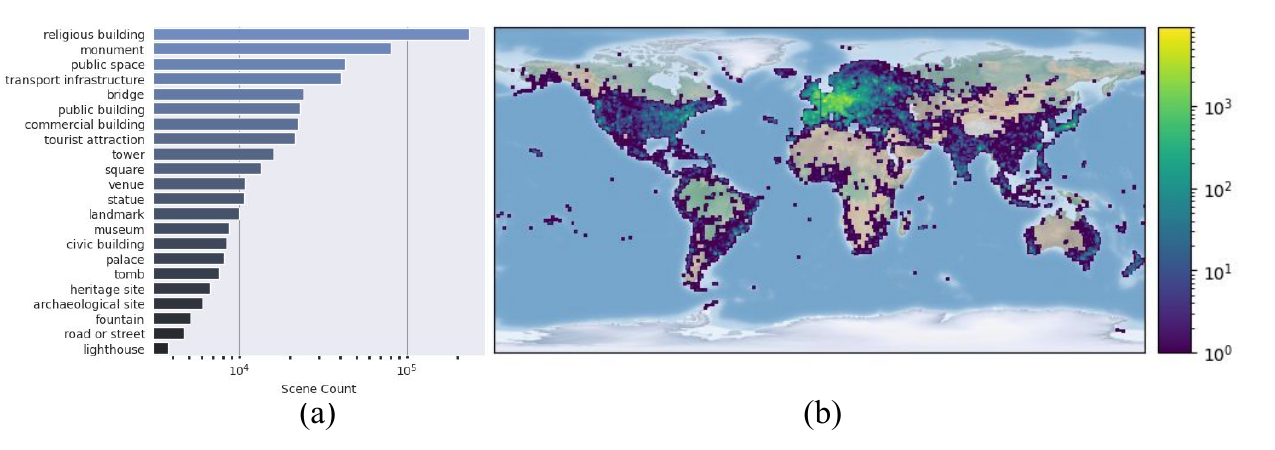}
    \caption{
    Distribution of the MegaScenes Dataset. 
    On the left, we depict the frequency of scenes grouped by Wikidata class. 
    This includes only select classes with more than 3,500 scenes; note that a single scene may be an instance of multiple classes. 
    On the right, we visualize the geospatial distribution of collected scenes worldwide.
    }
    \label{fig:scene_distrib}
    \vspace{-1.5em}
\end{figure}

\medskip
\noindent \textbf{Wikimedia Commons Categories as Scenes.}
We base each scene in MegaScenes from a single Wikimedia Commons category. Contributors from around the world have uploaded millions of images to Wikimedia Commons, and have organized images into representative groups. As shown in \cref{fig:scene_distrib} we find that Wikimedia Commons categories depict scenes that are distributed across Earth, making it suitable as the foundation for a diverse dataset and future expandability.

\medskip
\noindent \textbf{Images, Subcategorization, and Licensing.}
Images within a single scene are further classified into subcategories determined by Wikimedia Commons contributors. 
This enables future dataset applications to create subsets of data with greater granularity. This also proves to be helpful in cleaning the dataset, as described in the supplement on the dataset pipeline. 
Most importantly, like the similarly-sourced Google Landmarks v2 dataset~\cite{weyand2020GLDv2}, these images possess open content licenses or are in the public domain. Consequently, depending on the specific license, these images are free to reuse and alter for downstream tasks, so long as the original source is attributed.

\medskip
\noindent \textbf{3D Data.}
For each scene, we contribute SIFT~\cite{lowe2004distinctive} keypoints and descriptors, as well as calculated two-view geometries for pairs of images. We also contribute sparse point clouds and camera poses for a subset of scenes with ample image overlap. We use COLMAP~\cite{schoenberger2016sfm} to compute this data.

\medskip
\noindent \textbf{Class Hierarchy.}
Similar to hierarchical extension of Google Landmarks v2 \cite{ramzi2023optimization}, the MegaScenes Dataset contains a hierarchy of class labels for each scene directly sourced from Wikidata. Wikidata is a large database of structured data, ranging from singular scenes to broad ideas, connecting topics between Wikimedia Commons and Wikipedia.
We use this class hierarchy to aid in dataset curation, as described in \cref{sec:datacurate}. 

\vspace{-1em}
\subsection{Dataset Curation}
\label{sec:datacurate}
\cref{fig:dataset_pipeline} depicts our dataset curation pipeline, which has three main steps: identifying scene categories, downloading images, and reconstructing scenes. We provide an overview of our pipeline below, and supply additional details in the supplement.

\medskip
\noindent \textbf{Identifying Scenes.}
Our first goal is to identify Wikimedia Commons categories that may be considered as scenes. We take a top-down approach to identify scenes by utilizing the class hierachy described in \cref{sec:dataformat}, as follows. First, we select several broad classes from Wikidata, such as ``bridges'' or ``religious buildings'', that relate to collections of scenes. We choose these classes based on commonly seen places in everyday life. From these classes, we use the class hierarchy to identify Wikimedia Commons categories that are instances of these classes.

\medskip
\noindent \textbf{Downloading Images from Scenes.} We download all images associated with a Wikimedia Commons category that is identified as a scene, contingent on a subcategory filter we put in place to avoid downloading unrelated images. This filter is described in the supplement.

\medskip
\noindent \textbf{Reconstructing Scenes with SfM and Cleaning Reconstructions.}
For each scene, we run structure from motion on its corresponding collection of images using COLMAP~\cite{schoenberger2016sfm} to produce sparse point clouds and camera poses. We use default parameters for feature extraction, vocabulary tree matching \cite{schoenberger2016vote}, and sparse reconstruction.
We identify incorrect SfM reconstructions due to visual ambiguities (e.g., repeated patterns) by manual inspection guided by historically problematic scenes described in prior work~\cite{heinly2014_indistinguishable_geometry,cai2023doppelgangers}. 
For these scenes, we run the Doppelgangers \cite{cai2023doppelgangers} pipeline to get a corrected reconstruction.

\vspace{-1em}
\subsection{Dataset Statistics}
\label{sec:datastats}

In total, MegaScenes consists of approximately 430K scenes derived from Wikimedia Commons. Across these categories, we download 9M
images which results in over 30M
image pairs with estimated two-view geometries. 
Around 80K
of these scenes led to at least one sparse COLMAP reconstruction, resulting in over 100K
reconstructions and 2M
registered images.
In these sparse reconstructions we triangulate 400 million 3D points, with a mean track length of 5 images and a mean of 8,700 observations per registered image.  

Similar to Google Landmarks~\cite{weyand2020GLDv2}, MegaScenes has a wide range of scenes, with as many as 18K images to as few as zero per scene. As shown by \cref{fig:scene_distrib},
our scenes covers a diverse set of classes ranging from buildings and outdoor spaces, to statues and streets.

\vspace{-1em}
\section{MegaScenes Applied to Novel View Synthesis}
\label{sec:nvs}
\vspace{-0.5em}

In this section, we explore MegaScenes on a representative application: novel view synthesis (NVS) from a single image.
The goal is to take a reference image and generate a plausible image at a target pose that is consistent with the reference image.
Following Sargent \etal~\cite{sargent2023zeronvs}, we train and evaluate on image pairs with pseudo-ground-truth relative poses obtained via SfM. Note that we only consider the setting where the two views have overlapping visual content. In the supplement, we provide videos obtained through autoregressive generation. 

We start with testing state-of-the-art novel-view synthesis models, namely Zero-1-to-3~\cite{liu2023zero} and ZeroNVS~\cite{sargent2023zeronvs}, on MegaScenes, and demonstrate that these approaches fail to generalize to in-the-wild scenes. We then improve these models in two ways. First, simply fine-tuning these methods on large numbers of training pairs from MegaScenes leads to dramatically improved results on both Internet photos of scenes and three out-of-domain datasets. Second, we observe that these fine-tuned models still demonstrate inconsistencies between the requested pose and the synthesized image. We show that by adding an additional conditioning image where we approximately warp the input view into the target view, we improve pose consistency and novel view quality.

In \cref{sec:nvssetup}, we describe our setup. Then, we show results of finetuning baseline models on MegaScenes in \cref{sec:finetuning}. In \cref{sec:ourmethod}, we analyze failure cases of existing methods and propose our method to improve pose consistency. Finally, we evaluate our method on multiple datasets in \cref{sec:evalmegascenes} and \cref{sec:evaldomains}. 

\vspace{-1em}
\subsection{Setup: Data Mining and Evaluation}
\label{sec:nvssetup}
\begin{figure}[t]
    \centering
    \includegraphics[width=\textwidth]{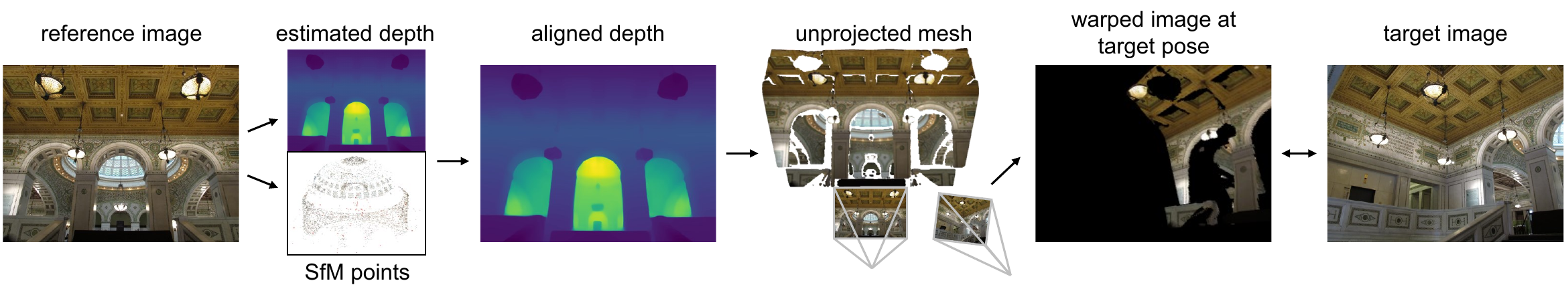}
    \caption{We create over 2 million pairs of training images for novel view synthesis. Each pair contains relative pose and a warping from the reference to target image which we use for both training and evaluation. We align estimated monocular depths with sparse point clouds from COLMAP~\cite{schoenberger2016sfm}, and unproject the RGBD images to a mesh for viewpoint rendering. See \cref{sec:nvssetup} for details and \cref{fig:megascenesfig} for more examples. }
    \label{fig:warp_figure}
\vspace{-1.5em}
\end{figure}

\noindent \textbf{Data Mining.}
We first identify a subset of image pairs from MegaScenes suitable for training novel view synthesis methods using two conditions. First, each pair should have similar lighting, since diffusion models operate on a pixel-wise loss. Using metadata, we find pairs of images taken within three hours of each other, as a proxy for lighting similarity. 
Second, we find pairs with sufficient visual overlap of at least 50 3D SfM points, so the model can learn view synthesis based on visual cues. 
With this threshold, we still observe both small and large pose changes, as shown in \cref{fig:megascenesfig}. Finally, we require that pairs have the same aspect ratio. Most previous works~\cite{rombach2022high, sargent2023zeronvs} center crop images, but we find that many landmarks, such as statues, can have highly varied aspect ratios and lose information through center cropping. Thus, we resize the long side to 256, and pad the short side, to obtain images with size 256$\times$256. 

As a final check, we manually inspect all scenes. We remove 298 scenes that we determine have too many occlusions in the majority of images; most of these occlusions were people.
In total, we obtain 2,086,036 pairs from 32,259 scenes and 475,277 unique images. We hold out 800 categories that contain 51,240 pairs and 11,852 unique images. We form our validation set from the first 10,000 pairs, which we use to determine model convergence. We form our test set with the remaining 41,240 pairs, which we use to report numbers.

\medskip
\noindent \textbf{Evaluation metrics.} 
We evaluate each method using standard reconstruction and generation metrics. 
For reconstruction, we calculate LPIPS~\cite{zhang2018perceptual}, PSNR, and SSIM~\cite{wang2004ssim}. LPIPS measures perceptual similarity, while PSNR and SSIM operate mainly on a pixel basis. However, generative models should only be expected to remain consistent with the target image where the pixels in the reference image are present, and retain the freedom to generate diverse samples, which could mean a lower reconstruction score. Thus, we propose ``masked'' versions of the metrics. We warp the input view into the target view using the target relative pose. Only pixels in the reference will be present in the warped image, and we only compare the copied pixels to the same location of the generated images. We create warpings by leveraging the sparse depth from 3D SfM points. First, we use Depth-Anything~\cite{depthanything} to estimate monocular depth of the reference image. We project the COLMAP sparse point cloud to the reference image's coordinate frame to obtain the ground-truth sparse depth, and use RANSAC to align the two. Now with the aligned dense depth, we unproject the reference RGBD to a mesh, and render it from the target pose. We visualize this process in \cref{fig:warp_figure}. 
For generative metrics, we use FID~\cite{heusel2017gansfid} and KID~\cite{binkowski2018demystifying}. Both assess the quality of generated images by comparing their feature distributions to  those of real images. Lower scores indicate that the generated images are more similar to real images.

In general, we find LPIPS, FID, and KID reliable metrics for assessing the \emph{quality} and \emph{realism} of generated images. We find Masked PSNR and Masked SSIM reliable for assessing \emph{consistency}, i.e. whether the generated images follow the target pose and retain details from the reference images. Still, we encourage readers to compare the qualitative results for a more comprehensive understanding of the properties of each model. 

\vspace{-1em}
\subsection{Finetuning Pose-Conditioned Models on MegaScenes}
\label{sec:finetuning}
Zero-1-to-3~\cite{liu2023zero} is finetuned from Stable Diffusion on Objaverse~\cite{objaverseXL}. ZeroNVS~\cite{sargent2023zeronvs} is finetuned from Zero-1-to-3 on CO3D~\cite{reizenstein21co3d}, ACID~\cite{infinite_nature_2020}, and RealEstate10K~\cite{RealEstate10k}. Our goal is to evaluate whether finetuning these models on MegaScenes improves generalization to in-the-wild scenes. 

\medskip
\noindent \textbf{Finetuning details.}
Zero-1-to-3 conditions on poses in spherical coordinates, which is only suitable for objects placed in a canonical coordinate frame, so for both models we follow ZeroNVS, which flattens the extrinsic matrix and field of view as input to cross-attention. The scale of translation is determined by the 20th quantile of the depth of the reference image~\cite{sargent2023zeronvs}. Additionally, both models concatenate the target and reference images and provide the CLIP~\cite{CLIPradford2021learning} embedding of the reference image to cross-attention so that the output remains consistent with the reference. Comparing the released and finetuned models verifies whether our dataset improves generalization to in-the-wild scenes. We provide training details in the supplement.

\begin{figure}[]
    \centering 
    \includegraphics[width=\textwidth]{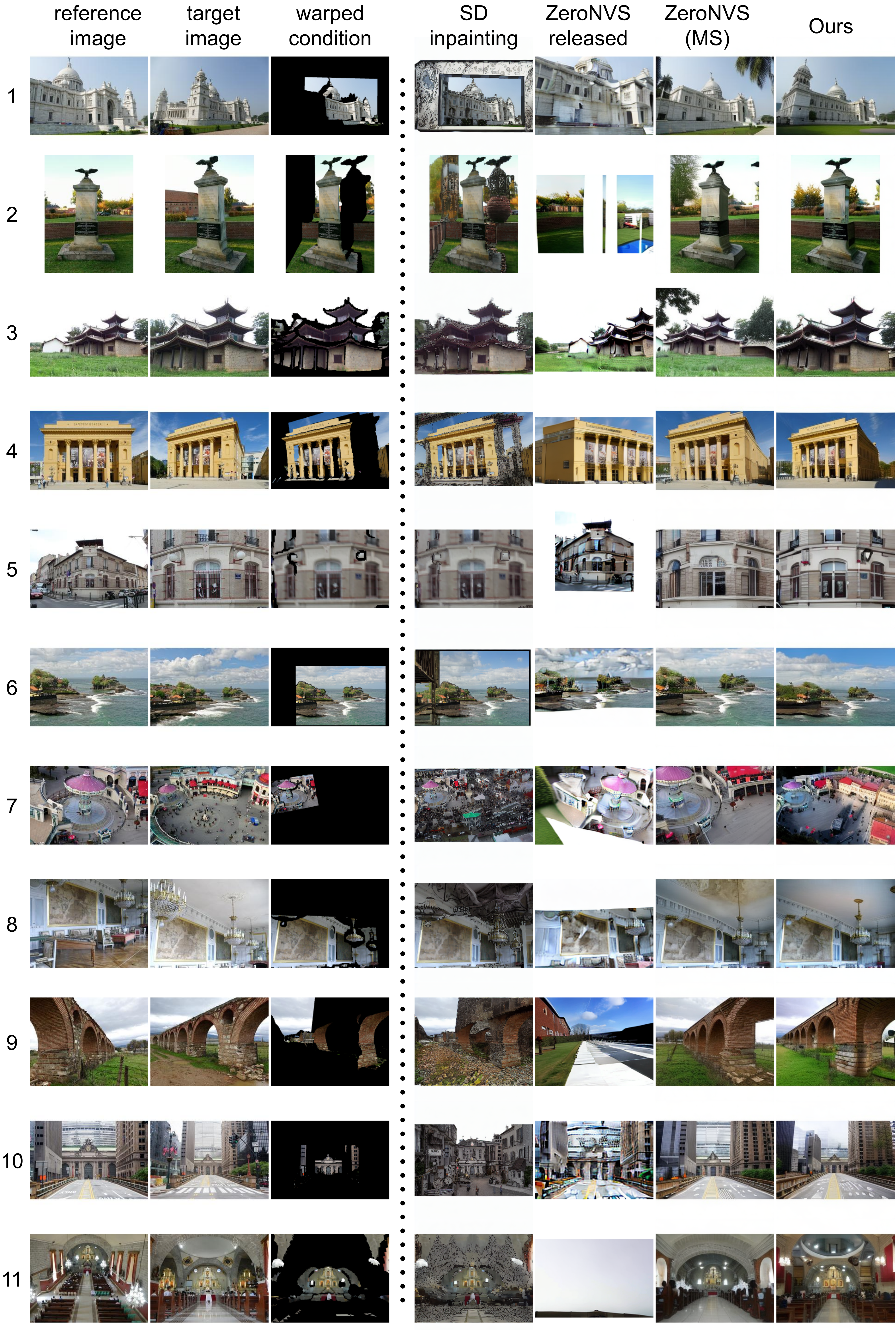}
    \caption{We evaluate multiple baselines on MegaScenes, which contains diverse scenes, poses, and object compositions. Prior methods exhibit many failure modes in this challenging setting. Our method identifies and addresses these failure modes.}
    \label{fig:megascenesfig}
\end{figure}

\medskip
\noindent \textbf{Results.}
We show qualitative results in \cref{fig:megascenesfig} and quantitative results in \cref{tab:internetphotos}. We denote the checkpoints released by authors with \textit{(released)} and the models finetuned on MegaScenes with \textit{(MS)}.
Due to space constraints, we only visualize the main baselines and models, but include the others in the supplement.

\textit{Zero-1-to-3 (released)} and \textit{ZeroNVS (released)} are both unable to generalize to internet photos. They produce unrealistic images with incorrect poses. We note that the former outperforms the latter in numbers, but upon inspecting qualitative results (see supplement) we observe that \textit{Zero-1-to-3 (released)} tends to return the reference image. Finetuning on MegaScenes signficantly improves results of both models, seen in the metrics of the \textit{(MS)} models and the qualitative comparisons between \textit{ZeroNVS (released)} and \textit{ZeroNVS (MS)}. 

We also validate that MegaScenes is suitable for the task of scene-level NVS. \textit{Zero-1-to-3 (MS)} outperforms \textit{ZeroNVS (released)} even though both are finetuned from Zero-1-to-3's released checkpoint; one is trained on MegaScenes and the other on CO3D~\cite{reizenstein21co3d}, ACID~\cite{infinite_nature_2020}, and RealEstate10K~\cite{RealEstate10k}.

\textit{ZeroNVS (MS)} shows the best performance among these four models. From \cref{fig:megascenesfig}, we see that it produces realistic images, and the generated images clearly attempt to follow the desired pose. 
However, many images produced by \textit{ZeroNVS (MS)} are still inaccurate. The positions of the islands (row 6), bridge (row 9), and building (row 10) are slightly different than in the target image, and when there is larger zoom, such as in rows 3, 4, and 7, the model fails to interpret the scale properly. Next, we address these issues.

\begin{table}[t]
    \caption{We evaluate whether models trained on MegaScenes generalize to in-the-wild scenes. \textit{Zero-1-to-3 / ZeroNVS (released)} are released checkpoints. We finetune both on MegaScenes (models denoted with \textit{(MS)}). 
    \textit{SD-inpainting} uses image warping and a pretrained diffusion inpainting model without finetuning, following the setup in \cite{yu2023wonderjourney, chung2023luciddreamer}. Our method takes warped images as input, and we condition with and without (\textit{w/o ext}) the extrinsic matrix. $\uparrow$ means higher is better and $\downarrow$ means lower is better.}
    \centering
    \resizebox{0.9\linewidth}{!} {%
      \begin{tabular}{lcccccccc}
    \toprule
     & LPIPS$(\downarrow)$ & PSNR$(\uparrow)$ & SSIM$(\uparrow)$& \makecell{Masked\\LPIPS }$(\downarrow)$ & \makecell{Masked\\PSNR} $(\uparrow)$ & \makecell{Masked\\SSIM} $(\uparrow)$ &FID$(\downarrow)$ & KID$(\downarrow)$ \\
    \midrule 
    \multicolumn{2}{l}{\textbf{\textit{Pose-Conditioned (\cref{sec:finetuning})}}}\\
    Zero-1-to-3 (released) & 0.5476 & 9.0896 & 0.2413 & 0.2777 & 14.132 & 0.6320 & 86.892 & 0.0634  \\
    ZeroNVS (released) &  0.6156 & 7.4711 & 0.1508 & 0.3229 & 11.041 & 0.5421 & 69.097 & 0.0487  \\
    Zero-1-to-3 (MS) & 0.4289 & 12.159 & 0.3665 & 0.1811 & 19.952 & 0.7286 & 9.7835 & 0.0023\\
    ZeroNVS (MS) & 0.3857 & 12.900 & 0.4005 & 0.1572 & 20.713 & 0.7534 & 9.8382 & 0.0024\\
    \multicolumn{2}{l}{\textbf{\textit{Warp-Conditioned (\cref{sec:ourmethod})}}}\\
    SD-inpainting & 0.4245 & 12.358 & 0.3923 & 0.1283 & 24.377 & 0.8005 & 38.484 & 0.0242\\
    Ours w/o ext & 0.3534 & 13.310 & 0.4328 & 0.1297 & 22.609 & 0.7819 & 12.010 & 0.0041\\
    \multicolumn{2}{l}{\textbf{\textit{Warp + Pose (\cref{sec:ourmethod})}}}\\
    Ours & 0.3444 & 13.397 & 0.4446 & 0.1256 & 22.483 & 0.7842 & 11.580 & 0.0040\\
    \bottomrule
    \end{tabular} %
    }
    \label{tab:internetphotos}
    \vspace{-1em}
\end{table}

\vspace{-1em}
\subsection{Improving Pose Consistency with Warp Conditioning}
\label{sec:ourmethod}
ZeroNVS~\cite{sargent2023zeronvs} conditions its diffusion model on the flattened extrinsic matrix, which is not a very intuitive pose condition; the model needs to learn the spatial transformation without visual cues.
Furthermore, the translation scale is ambiguous since scenes cannot be canonicalized to a fixed coordinate frame. In the original paper, the authors run a grid search on each scene to manually determine a scene scale during inference. Since the goal is to generalize to in-the-wild scenes, when testing ZeroNVS we determine scene scale automatically based on monocular depth estimation, which leads to inaccurate poses especially when there are large zooming effects. 

Our insight is that the warped image (\cref{fig:warp_figure}) encodes pose by visualizing how pixels are supposed to move, and is directly aligned with the scene scale. 
On our training and evaluation datasets, the scale is based on 3D SfM points.
Given a single image, we can determine the scene scale from estimated monocular depth and use the same extrinsics for conditioning and warping the image for a consistent scale. We show this setting in the supplement by generating videos given a single image. Thus, we concatenate the warped image with the input target and reference images, and observe significant improvements in pose accuracy. 

However, using only the warped image as pose condition leads to two problems. 
First, inaccurate depth, which can be common in noisy, in-the-wild scenes, can cause the model to fail. Furthermore, with guidance only through the movement of 2D pixels, the model seems unable to interpret 3D structure at times. 
\begin{wrapfigure}{r}{0.5\textwidth} %
\vspace{-1.5em}
\includegraphics[width=0.5\textwidth]{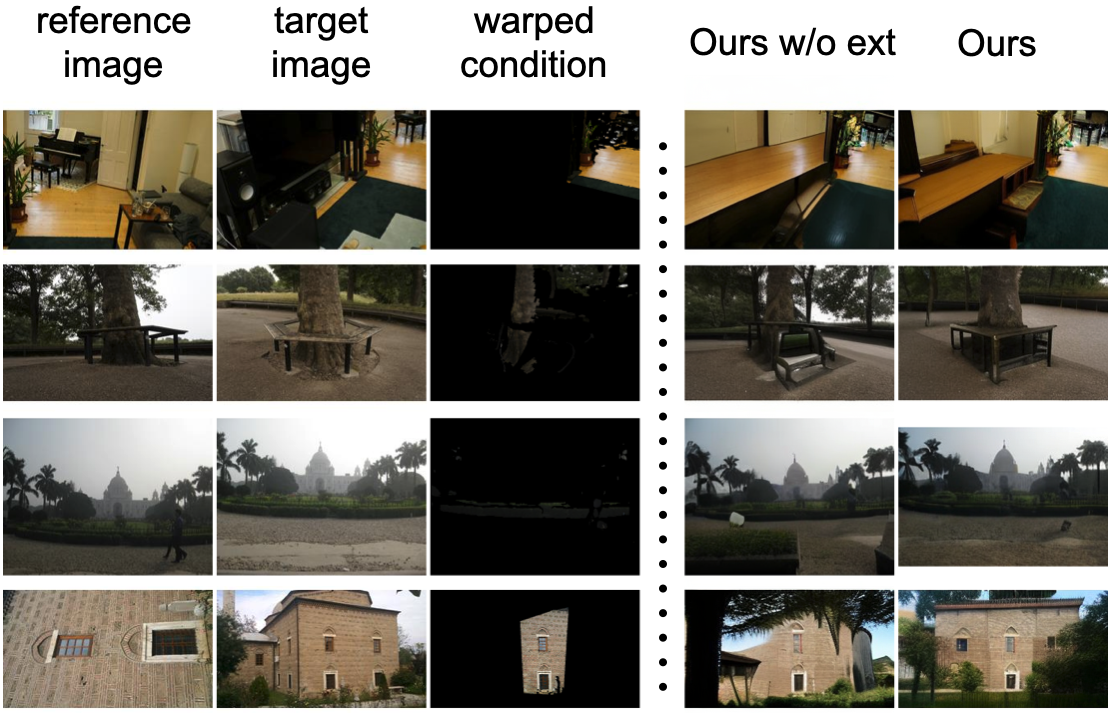}
 \caption{We compare results with and without conditioning on the extrinsic matrix. The extrinsic matrix ensures valid outputs and produces more consistent 3D geometry.}
\label{fig:extablation}
\vspace{-2em}
\end{wrapfigure}
Therefore, we also condition on the extrinsic matrix following ZeroNVS. We show qualitative results of these design choices in \cref{fig:extablation}. In rows 1 and 4, \textit{Ours} demonstrate better 3D consistency compared to \textit{Ours w/o ext}, which is the model trained without conditioning on the extrinsic matrix, including creating the separating wall and a complete building, respectively. In rows 2 and 3, there is little information in the warped condition due to inaccurate depth, and the generated results contain either unwanted objects or objects at inaccurate locations. 

Using the warped image as a condition for view synthesis was first proposed by Liu \etal~\cite{infinite_nature_2020}, then adapted by a recent line of work that uses a Stable Diffusion inpainting model without any finetuning~\cite{yu2023wonderjourney, chung2023luciddreamer}. The premise is that a foundation model can generalize to any domain without forgetting.
We therefore also compare to this baseline, denoted \textit{SD-inpainting}. Using the preprocessed warpings, we set empty pixels as the mask to inpaint. This method, however, demonstrates lack of 3D understanding. For instance, the inpainting model frequently interprets a scene as a picture frame. Furthermore, the inpainting model is only trained on large, uniform masks, and produces artifacts when operating on fine-grained masks, which come with arbitrary poses. In contrast, the masks can be of arbitrary size, and our diffusion model has the freedom to remove occlusions and create plausible images. These results suggest that finetuning on 3D data is essential for zero-shot novel view synthesis.

In the following subsections, we evaluate all methods on MegaScenes as well as three out-of-domain datasets. While our method is simple and builds on existing methods, it addresses the fundamental issues of prior works, and we validate that it produces significantly more consistent and realistic results. We show a large collection of uncurated results in the supplement and demonstrate that our method is effective across a variety of diverse scenes.

\begin{table}[t!]
    \caption{We evaluate whether models trained on MegaScenes generalize to other data domains. The models and metrics are the same as in \cref{tab:internetphotos}. We test on DTU, Mip-NeRF360, and RealEstate10K. }
    \centering
    \resizebox{0.9\linewidth}{!} {
      \begin{tabular}{lcccccccc}
    \toprule
     \textbf{DTU} & LPIPS$(\downarrow)$ & PSNR$(\uparrow)$ & SSIM$(\uparrow)$& \makecell{Masked\\LPIPS}$(\downarrow)$ & \makecell{Masked\\PSNR}$(\uparrow)$ & \makecell{Masked\\SSIM}$(\uparrow)$ &FID$(\downarrow)$ & KID$(\downarrow)$ \\
    \midrule 
    \multicolumn{2}{l}{\textbf{\textit{Pose-Conditioned (\cref{sec:finetuning})}}}\\
    Zero-1-to-3 (released) & 0.5647 & 6.8720 & 0.2100 & 0.2592 & 12.628 & 0.6609 & 128.93 & 0.0297   \\
    ZeroNVS (released) & 0.6476 & 5.7992 & 0.1113 & 0.3193 & 9.7005 & 0.5517 & 159.96 & 0.0352   \\
    Zero-1-to-3 (MS) & 0.5158 & 7.6367 & 0.2755 & 0.2080 & 13.311 & 0.7014 & 101.94 & 0.0223 \\
    ZeroNVS (MS) & 0.4833 & 8.0191 & 0.3066 & 0.1908 & 13.515 & 0.7152 & 87.406 & 0.0158\\
    \multicolumn{2}{l}{\textbf{\textit{Warp-Conditioned (\cref{sec:ourmethod})}}}\\
    SD-inpainting & 0.4951 & 9.9463 & 0.3688 & 0.1283 & 22.656 & 0.8333 & 214.42 & 0.1067 \\
    Ours w/o ext & 0.4113 & 8.8473 & 0.3878 & 0.1385 & 16.631 & 0.7924 & 92.284 & 0.0193\\
    \multicolumn{2}{l}{\textbf{\textit{Warp + Pose (\cref{sec:ourmethod})}}}\\
    Ours & 0.3995 & 8.7953 & 0.3930 & 0.1357 & 16.593 & 0.7916 & 85.959 & 0.0163 \\
    \bottomrule 
    \end{tabular}}
    \vspace{1em}
    
    \resizebox{0.9\linewidth}{!} {
    \begin{tabular}{lcccccccc}
    \toprule 
    \textbf{Mip-NeRF360} & LPIPS$(\downarrow)$ & PSNR$(\uparrow)$ & SSIM$(\uparrow)$& \makecell{Masked\\LPIPS}$(\downarrow)$ & \makecell{Masked\\PSNR}$(\uparrow)$ & \makecell{Masked\\SSIM}$(\uparrow)$ &FID$(\downarrow)$ & KID$(\downarrow)$ \\
    \midrule 
    \textbf{\textit{Pose-Conditioned}}\\
    Zero-1-to-3 (released) & 0.5258 & 10.720 & 0.2865 & 0.1621 & 16.299 & 0.8864 & 171.21 & 0.1126   \\
    ZeroNVS (released) &  0.6685 & 6.9993 & 0.1240 & 0.2312 & 10.890 & 0.7670 & 137.04 & 0.0537  \\
    Zero-1-to-3 (MS) & 0.4429 & 12.921 & 0.3828 & 0.0307 & 29.441 & 0.9697 & 67.645 & 0.0163\\
    ZeroNVS (MS) &0.4057 & 13.780 & 0.4122 & 0.1369 & 24.909 & 0.8219 & 60.677 & 0.0139 \\
    \textbf{\textit{Warp-Conditioned}}&\\
    SD-inpainting & 0.4557 & 12.922 & 0.3996 & 0.1212 & 27.455 & 0.8488 & 150.11 & 0.0792\\
    Ours w/o ext & 0.3944 & 13.667 & 0.4279 & 0.1237 & 25.884 & 0.8344 & 70.684 & 0.0193 \\
    \textbf{\textit{Warp + Pose}}& \\
    Ours & 0.3807 & 14.056 & 0.4406 & 0.1150 & 26.196 & 0.8422 & 64.406 & 0.0142\\
    \bottomrule 
    \end{tabular}}
    \vspace{1em}
    
    \resizebox{0.9\linewidth}{!} {
    \begin{tabular}{lcccccccc}
    \toprule 
    \textbf{RE10K} & LPIPS$(\downarrow)$ & PSNR$(\uparrow)$ & SSIM$(\uparrow)$& \makecell{Masked\\LPIPS}$(\downarrow)$ & \makecell{Masked\\PSNR}$(\uparrow)$ & \makecell{Masked\\SSIM}$(\uparrow)$ &FID$(\downarrow)$ & KID$(\downarrow)$ \\
    \midrule
    \textbf{\textit{Pose-Conditioned}}\\
    Zero-1-to-3 (released) &  0.4050 & 11.632 & 0.4384 & 0.2732 & 14.079 & 0.6400 & 160.20 & 0.0725  \\
    ZeroNVS (released) &  0.4563 & 9.4869 & 0.3527 & 0.3078 & 11.456 & 0.5565 & 123.01 & 0.0352   \\
    Zero-1-to-3 (MS) & 0.2722 & 14.638 & 0.5697 & 0.1510 & 21.241 & 0.7637 & 68.908 & 0.0024 \\
    ZeroNVS (MS) & 0.2053 & 16.015 & 0.6304 & 0.1176 & 20.609 & 0.8070 & 61.117 & 0.0024\\
    \textbf{\textit{Warp-Conditioned}}&\\
    SD-inpainting & 0.2694 & 15.541 & 0.6429 & 0.0929 & 29.056 & 0.8719 & 118.94 & 0.0396\\
    Ours w/o ext & 0.1922 & 16.105 & 0.6267 & 0.1109 & 23.147 & 0.7985 & 66.770 & 0.0057\\
    \textbf{\textit{Warp + Pose}}& \\
    Ours & 0.1774 & 17.224 & 0.6661 & 0.0942 & 24.259 & 0.8315 & 60.013 & 0.0023 \\
    \bottomrule
    \end{tabular} %
    }
    \label{tab:diffdomains}
    \vspace{-1em}
\end{table}

\vspace{-1em}
\subsection{Evaluation on MegaScenes}
\label{sec:evalmegascenes}

\begin{figure}[]
    \centering 
    \includegraphics[width=\textwidth]{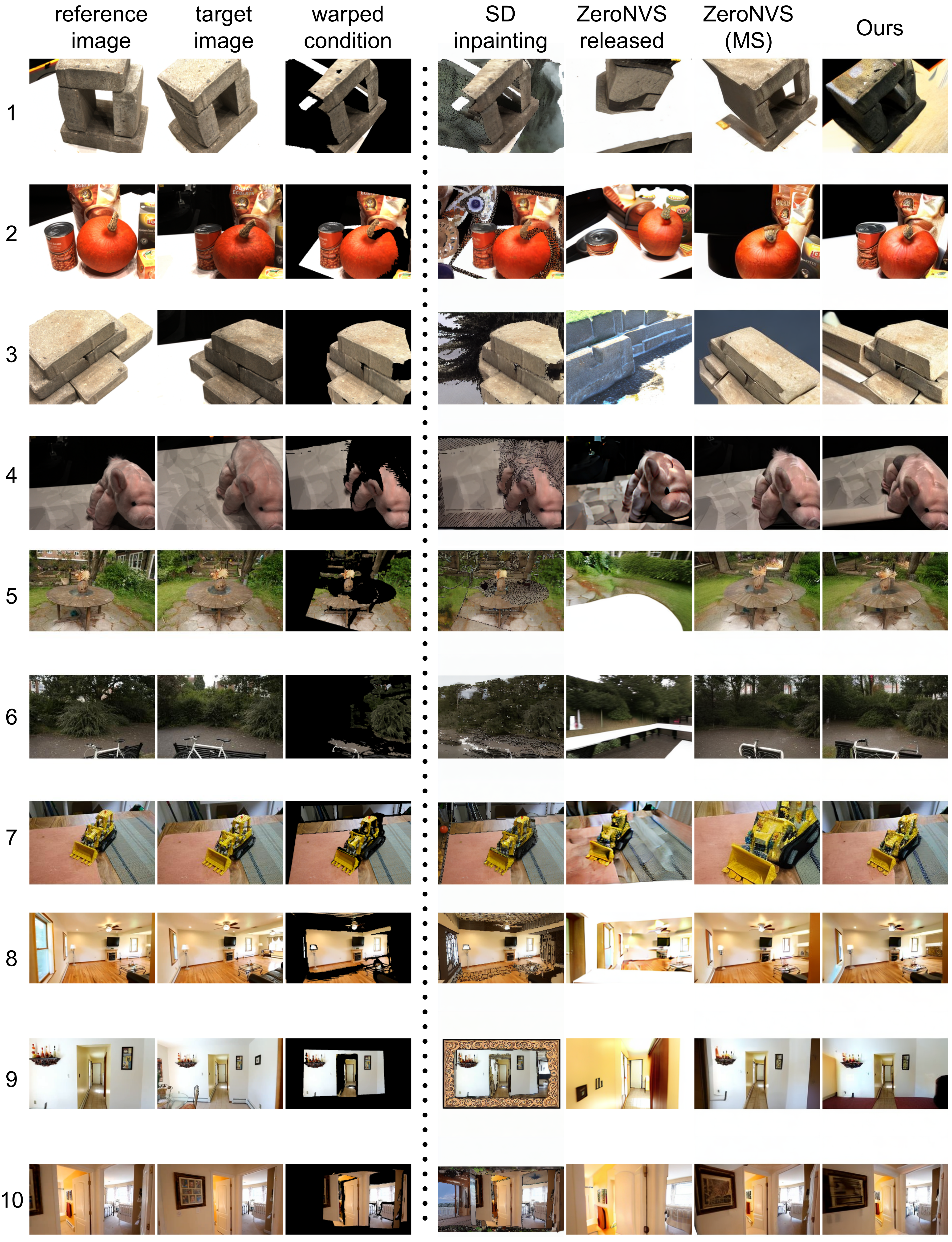}
    \caption{We evaluate multiple models on DTU (rows 1-4), MipNerf-360 (5-7), and Re10K (8-10). Models trained on MegaScenes are able to generalize to these datasets.
    }
    \label{fig:domainsfig}
\end{figure}

We first evaluate on MegaScenes' test set, which consists of in-the-wild scenes from Internet photos. We show quantitative results in \cref{tab:internetphotos} and qualitative results in \cref{fig:megascenesfig}. 
Our method produces images closest to the desired pose, while being realistic and visually consistent with the reference. Compared to \textit{ZeroNVS (MS)}, our method places objects according to cues in the warped image, which not only leads to more accurate positioning (and better reconstruction metrics), but also more detail (structure in row 5, statues at the end of the hall in row 11). However, our method also has higher FID and KID scores compared to the pose-conditioned models because the warped images add constraint to the generation process. We do not visually observe a degradation in image quality though.

\textit{SD-inpainting} has the best reconstruction metrics, as it faithfully returns the pixels in the warped condition. However, it does not understand 3D geometry, evident in the inconsistent generations, such as interpreting a scene as a picture frame (row 1). The inpainting artifacts also lead to unrealistic images, resulting in high LPIPS, FID, and KID scores. Our method avoids these issues while taking advantage of strong position cues of the warped condition. 

\vspace{-1em}
\subsection{Evaluation on Datasets from Different Domains}
\label{sec:evaldomains}

Next, we evaluate on DTU~\cite{dtu}, Mip-NeRF 360~\cite{barron2022mipnerf360}, and RealEstate10K (Re10K)~\cite{RealEstate10k}, datasets commonly used for evaluating novel view synthesis. 

We expect MegaScenes to be sufficiently diverse such that models trained on it can generalize even to specific domains. We obtain image pairs and warpings from all three datasets. In total, we obtain 2,850 evaluation pairs from DTU, 15,682 pairs from Mip-NeRF 360, and 644 pairs from Re10K. We describe our data setup in the supplement.

We show results in \cref{tab:diffdomains} and \cref{fig:domainsfig}. We see a similar trend as the previous subsection. All metrics and qualitative results improve when trained on MegaScenes. \textit{ZeroNVS (released)} was trained on object-centric datasets similar to DTU and MipNeRF 360, and directly trained on Re10K, but \textit{Zero-1-to-3 (MS)} signficantly outperforms it on all datasets. This validates that MegaScenes' categories cover a wide variety of domains.

Again, our method produces images closest to the desired pose. \textit{ZeroNVS (MS)} mostly follows the pose condition, but the positions of the objects are less accurate. This is obvious in DTU where we can visually match the corners of the objects. Direct visual cues in the warped condition allow our model to preserve structure even in challenging cases, such as the bicycle in row 6. 

As a side note, we would like to point out that \textit{ZeroNVS (released)} appears to perform worse than shown in their original paper because of different testing settings. The original paper evaluates results only after SDS~\cite{poole2022dreamfusion} optimization, which filters out noise and samples the mode of the diffusion outputs. Additionally, the authors run a grid search to manually determine scene scale. We note that we were able to reproduce results shown in the original paper, but here we present feed-forward results without optimization or manual tuning for fair comparison. 

\vspace{-1em}
\section{Conclusion}
\vspace{-0.5em}
We present MegaScenes, a general large-scale 3D dataset, and analyze its impact on scene-level novel view synthesis. We find that finetuning NVS methods on MegaScenes significantly improves synthesis quality, validating the coverage of the dataset. We also improve existing methods and observe increased pose accuracy.

\textbf{Limitations and Future Work.} On the task of NVS, we use a fraction of our data (475K out of 2M images) and a subset of data types (we did not use text captions). We would like to expand MegaScenes to applications that leverage the full dataset. Our NVS method also comes with limitations. It relies on warped images for conditioning and is impacted by erroneous depth estimation. Also, it cannot handle large camera motions such as behind a scene. Finally, we bypass lighting by sampling based on metadata, but we could incorporate lighting conditions~\cite{li2020crowdsampling, martinbrualla2020nerfw} in the future.

\section*{Acknowledgments}
We thank Brandon Li for building the COLMAP webviewer. 
This work was funded in part by the National Science Foundation (IIS-2008313, IIS-2211259, IIS-2212084). Gene Chou was funded by a NSF Graduate Research Fellowship.

\bibliographystyle{splncs04}
\bibliography{main}

\begin{thebibliography}{10}
\providecommand{\url}[1]{\texttt{#1}}
\providecommand{\urlprefix}{URL }
\providecommand{\doi}[1]{https://doi.org/#1}

\bibitem{barron2022mipnerf360}
Barron, J.T., Mildenhall, B., Verbin, D., Srinivasan, P.P., Hedman, P.: Mip-nerf 360: Unbounded anti-aliased neural radiance fields. CVPR  (2022)

\bibitem{zoedepth}
Bhat, S.F., Birkl, R., Wofk, D., Wonka, P., Müller, M.: Zoedepth: Zero-shot transfer by combining relative and metric depth (2023). \doi{10.48550/ARXIV.2302.12288}, \url{https://arxiv.org/abs/2302.12288}

\bibitem{binkowski2018demystifying}
Bi{\'n}kowski, M., Sutherland, D.J., Arbel, M., Gretton, A.: Demystifying mmd gans. arXiv preprint arXiv:1801.01401  (2018)

\bibitem{cai2023doppelgangers}
Cai, R., Tung, J., Wang, Q., Averbuch-Elor, H., Hariharan, B., Snavely, N.: Doppelgangers: Learning to disambiguate images of similar structures. In: ICCV (2023)

\bibitem{cai2022diffdreamer}
Cai, S., Chan, E.R., Peng, S., Shahbazi, M., Obukhov, A., Van~Gool, L., Wetzstein, G.: Diffdreamer: Towards consistent unsupervised single-view scene extrapolation with conditional diffusion models. In: ICCV (2023)

\bibitem{chan2023generative}
Chan, E.R., Nagano, K., Chan, M.A., Bergman, A.W., Park, J.J., Levy, A., Aittala, M., De~Mello, S., Karras, T., Wetzstein, G.: Generative novel view synthesis with 3d-aware diffusion models. arXiv preprint arXiv:2304.02602  (2023)

\bibitem{chang2017matterport3d}
Chang, A., Dai, A., Funkhouser, T., Halber, M., Niessner, M., Savva, M., Song, S., Zeng, A., Zhang, Y.: Matterport3d: Learning from rgb-d data in indoor environments. arXiv preprint arXiv:1709.06158  (2017)

\bibitem{shapenet}
Chang, A.X., Funkhouser, T., Guibas, L., Hanrahan, P., Huang, Q., Li, Z., Savarese, S., Savva, M., Song, S., Su, H., et~al.: Shapenet: An information-rich 3d model repository. arXiv preprint arXiv:1512.03012  (2015)

\bibitem{chung2023luciddreamer}
Chung, J., Lee, S., Nam, H., Lee, J., Lee, K.M.: Luciddreamer: Domain-free generation of 3d gaussian splatting scenes. arXiv preprint arXiv:2311.13384  (2023)

\bibitem{manhattanworld}
Coughlan, J.M., Yuille, A.L.: Manhattan world: Compass direction from a single image by bayesian inference. In: ICCV (1999)

\bibitem{dai2017scannet}
Dai, A., Chang, A.X., Savva, M., Halber, M., Funkhouser, T., Nie{\ss}ner, M.: Scannet: Richly-annotated 3d reconstructions of indoor scenes. In: Proceedings of the IEEE conference on computer vision and pattern recognition. pp. 5828--5839 (2017)

\bibitem{objaverseXL}
Deitke, M., Liu, R., Wallingford, M., Ngo, H., Michel, O., Kusupati, A., Fan, A., Laforte, C., Voleti, V., Gadre, S.Y., VanderBilt, E., Kembhavi, A., Vondrick, C., Gkioxari, G., Ehsani, K., Schmidt, L., Farhadi, A.: Objaverse-xl: A universe of 10m+ 3d objects. arXiv preprint arXiv:2307.05663  (2023)

\bibitem{du2021nerflow}
Du, Y., Zhang, Y., Yu, H.X., Tenenbaum, J.B., Wu, J.: Neural radiance flow for 4d view synthesis and video processing. In: Proceedings of the IEEE/CVF International Conference on Computer Vision (2021)

\bibitem{edstedt2024dedode}
Edstedt, J., Bökman, G., Wadenbäck, M., Felsberg, M.: {DeDoDe: Detect, Don't Describe --- Describe, Don't Detect for Local Feature Matching}. In: 2024 International Conference on 3D Vision (3DV). IEEE (2024)

\bibitem{heinly2014_indistinguishable_geometry}
Heinly, J., Dunn, E., Frahm, J.M.: {Recovering Correct Reconstructions from Indistinguishable Geometry}. In: International Conference on 3D Vision (3DV) (2014)

\bibitem{heusel2017gansfid}
Heusel, M., Ramsauer, H., Unterthiner, T., Nessler, B., Hochreiter, S.: Gans trained by a two time-scale update rule converge to a local nash equilibrium. Advances in neural information processing systems  \textbf{30} (2017)

\bibitem{ho2022classifier}
Ho, J., Salimans, T.: Classifier-free diffusion guidance. arXiv preprint arXiv:2207.12598  (2022)

\bibitem{hong2023lrm}
Hong, Y., Zhang, K., Gu, J., Bi, S., Zhou, Y., Liu, D., Liu, F., Sunkavalli, K., Bui, T., Tan, H.: Lrm: Large reconstruction model for single image to 3d. arXiv preprint arXiv:2311.04400  (2023)

\bibitem{invs2023}
Kant, Y., Siarohin, A., Vasilkovsky, M., Guler, R.A., Ren, J., Tulyakov, S., Gilitschenski, I.: invs : Repurposing diffusion inpainters for novel view synthesis. In: SIGGRAPH Asia 2023 Conference Papers (2023)

\bibitem{kerbl20233d}
Kerbl, B., Kopanas, G., Leimk{\"u}hler, T., Drettakis, G.: 3d gaussian splatting for real-time radiance field rendering. ACM Transactions on Graphics  \textbf{42}(4) (2023)

\bibitem{li2023matrixcity}
Li, Y., Jiang, L., Xu, L., Xiangli, Y., Wang, Z., Lin, D., Dai, B.: Matrixcity: A large-scale city dataset for city-scale neural rendering and beyond. arXiv e-prints pp. arXiv--2308 (2023)

\bibitem{MegaDepthLi18}
Li, Z., Snavely, N.: Megadepth: Learning single-view depth prediction from internet photos. In: Computer Vision and Pattern Recognition (CVPR) (2018)

\bibitem{li2020crowdsampling}
Li, Z., Xian, W., Davis, A., Snavely, N.: Crowdsampling the plenoptic function. In: European Conference on Computer Vision. pp. 178--196. Springer (2020)

\bibitem{li2021openrooms}
Li, Z., Yu, T.W., Sang, S., Wang, S., Song, M., Liu, Y., Yeh, Y.Y., Zhu, R., Gundavarapu, N., Shi, J., et~al.: Openrooms: An open framework for photorealistic indoor scene datasets. In: Proceedings of the IEEE/CVF conference on computer vision and pattern recognition. pp. 7190--7199 (2021)

\bibitem{lindenberger2023lightglue}
Lindenberger, P., Sarlin, P.E., Pollefeys, M.: {LightGlue: Local Feature Matching at Light Speed}. In: ICCV (2023)

\bibitem{ling2023dl3dv}
Ling, L., Sheng, Y., Tu, Z., Zhao, W., Xin, C., Wan, K., Yu, L., Guo, Q., Yu, Z., Lu, Y., et~al.: Dl3dv-10k: A large-scale scene dataset for deep learning-based 3d vision. arXiv preprint arXiv:2312.16256  (2023)

\bibitem{infinite_nature_2020}
Liu, A., Tucker, R., Jampani, V., Makadia, A., Snavely, N., Kanazawa, A.: Infinite nature: Perpetual view generation of natural scenes from a single image. In: Proceedings of the IEEE/CVF International Conference on Computer Vision (ICCV) (October 2021)

\bibitem{liu2024one}
Liu, M., Xu, C., Jin, H., Chen, L., Varma~T, M., Xu, Z., Su, H.: One-2-3-45: Any single image to 3d mesh in 45 seconds without per-shape optimization. Advances in Neural Information Processing Systems  \textbf{36} (2024)

\bibitem{liu2023zero}
Liu, R., Wu, R., Van~Hoorick, B., Tokmakov, P., Zakharov, S., Vondrick, C.: Zero-1-to-3: Zero-shot one image to 3d object. In: Proceedings of the IEEE/CVF International Conference on Computer Vision. pp. 9298--9309 (2023)

\bibitem{liu2023syncdreamer}
Liu, Y., Lin, C., Zeng, Z., Long, X., Liu, L., Komura, T., Wang, W.: Syncdreamer: Generating multiview-consistent images from a single-view image. arXiv preprint arXiv:2309.03453  (2023)

\bibitem{lowe2004distinctive}
Lowe, D.G.: Distinctive image features from scale-invariant keypoints. Int. J. Comput. Vis.  \textbf{60}(2),  91--110 (2004)

\bibitem{martinbrualla2020nerfw}
Martin-Brualla, R., Radwan, N., Sajjadi, M.S.M., Barron, J.T., Dosovitskiy, A., Duckworth, D.: {NeRF in the Wild: Neural Radiance Fields for Unconstrained Photo Collections}. In: CVPR (2021)

\bibitem{mildenhall2021nerf}
Mildenhall, B., Srinivasan, P.P., Tancik, M., Barron, J.T., Ramamoorthi, R., Ng, R.: Nerf: Representing scenes as neural radiance fields for view synthesis. Communications of the ACM  \textbf{65}(1),  99--106 (2021)

\bibitem{niemeyer2022regnerf}
Niemeyer, M., Barron, J.T., Mildenhall, B., Sajjadi, M.S., Geiger, A., Radwan, N.: Regnerf: Regularizing neural radiance fields for view synthesis from sparse inputs. In: Proceedings of the IEEE/CVF Conference on Computer Vision and Pattern Recognition. pp. 5480--5490 (2022)

\bibitem{poole2022dreamfusion}
Poole, B., Jain, A., Barron, J.T., Mildenhall, B.: Dreamfusion: Text-to-3d using 2d diffusion. arXiv preprint arXiv:2209.14988  (2022)

\bibitem{qian2023magic123}
Qian, G., Mai, J., Hamdi, A., Ren, J., Siarohin, A., Li, B., Lee, H.Y., Skorokhodov, I., Wonka, P., Tulyakov, S., et~al.: Magic123: One image to high-quality 3d object generation using both 2d and 3d diffusion priors. arXiv preprint arXiv:2306.17843  (2023)

\bibitem{CLIPradford2021learning}
Radford, A., Kim, J.W., Hallacy, C., Ramesh, A., Goh, G., Agarwal, S., Sastry, G., Askell, A., Mishkin, P., Clark, J., et~al.: Learning transferable visual models from natural language supervision. In: International conference on machine learning. pp. 8748--8763. PMLR (2021)

\bibitem{ramzi2023optimization}
Ramzi, E., Audebert, N., Rambour, C., Araujo, A., Bitot, X., Thome, N.: {Optimization of Rank Losses for Image Retrieval}. In: In submission to: IEEE Transactions on Pattern Analysis and Machine Intelligence (2023)

\bibitem{reizenstein21co3d}
Reizenstein, J., Shapovalov, R., Henzler, P., Sbordone, L., Labatut, P., Novotny, D.: Common objects in 3d: Large-scale learning and evaluation of real-life 3d category reconstruction. In: International Conference on Computer Vision (2021)

\bibitem{roberts2021hypersim}
Roberts, M., Ramapuram, J., Ranjan, A., Kumar, A., Bautista, M.A., Paczan, N., Webb, R., Susskind, J.M.: Hypersim: A photorealistic synthetic dataset for holistic indoor scene understanding. In: Proceedings of the IEEE/CVF international conference on computer vision. pp. 10912--10922 (2021)

\bibitem{rombach2022high}
Rombach, R., Blattmann, A., Lorenz, D., Esser, P., Ommer, B.: High-resolution image synthesis with latent diffusion models. In: Proceedings of the IEEE/CVF conference on computer vision and pattern recognition. pp. 10684--10695 (2022)

\bibitem{saharia2022photorealistic}
Saharia, C., Chan, W., Saxena, S., Li, L., Whang, J., Denton, E.L., Ghasemipour, K., Gontijo~Lopes, R., Karagol~Ayan, B., Salimans, T., et~al.: Photorealistic text-to-image diffusion models with deep language understanding. Advances in Neural Information Processing Systems  \textbf{35},  36479--36494 (2022)

\bibitem{sargent2023zeronvs}
Sargent, K., Li, Z., Shah, T., Herrmann, C., Yu, H.X., Zhang, Y., Chan, E.R., Lagun, D., Fei-Fei, L., Sun, D., et~al.: Zeronvs: Zero-shot 360-degree view synthesis from a single real image. arXiv preprint arXiv:2310.17994  (2023)

\bibitem{schoenberger2016sfm}
Sch\"{o}nberger, J.L., Frahm, J.M.: Structure-from-motion revisited. In: Conference on Computer Vision and Pattern Recognition (CVPR) (2016)

\bibitem{schoenberger2016vote}
Sch\"{o}nberger, J.L., Price, T., Sattler, T., Frahm, J.M., Pollefeys, M.: A vote-and-verify strategy for fast spatial verification in image retrieval. In: Asian Conference on Computer Vision (ACCV) (2016)

\bibitem{shi2023mvdream}
Shi, Y., Wang, P., Ye, J., Long, M., Li, K., Yang, X.: Mvdream: Multi-view diffusion for 3d generation. arXiv preprint arXiv:2308.16512  (2023)

\bibitem{DDIM}
Song, J., Meng, C., Ermon, S.: Denoising diffusion implicit models. arXiv preprint arXiv:2010.02502  (2020)

\bibitem{dtu}
Sølund, T., Buch, A.G., Krüger, N., Aanæs, H.: A large scale 3d object recognition dataset. In: 3DV (2016)

\bibitem{tewari2024diffusion}
Tewari, A., Yin, T., Cazenavette, G., Rezchikov, S., Tenenbaum, J., Durand, F., Freeman, B., Sitzmann, V.: Diffusion with forward models: Solving stochastic inverse problems without direct supervision. Advances in Neural Information Processing Systems  \textbf{36} (2024)

\bibitem{tyszkiewicz2020disk}
Tyszkiewicz, M., Fua, P., Trulls, E.: Disk: Learning local features with policy gradient. Advances in Neural Information Processing Systems  \textbf{33} (2020)

\bibitem{verbin2022ref}
Verbin, D., Hedman, P., Mildenhall, B., Zickler, T., Barron, J.T., Srinivasan, P.P.: Ref-nerf: Structured view-dependent appearance for neural radiance fields. In: 2022 IEEE/CVF Conference on Computer Vision and Pattern Recognition (CVPR). pp. 5481--5490. IEEE (2022)

\bibitem{wang2023score}
Wang, H., Du, X., Li, J., Yeh, R.A., Shakhnarovich, G.: Score jacobian chaining: Lifting pretrained 2d diffusion models for 3d generation. In: Proceedings of the IEEE/CVF Conference on Computer Vision and Pattern Recognition. pp. 12619--12629 (2023)

\bibitem{wang2023pd}
Wang, J., Rupprecht, C., Novotny, D.: {PoseDiffusion}: Solving pose estimation via diffusion-aided bundle adjustment (2023)

\bibitem{wang2023pf}
Wang, P., Tan, H., Bi, S., Xu, Y., Luan, F., Sunkavalli, K., Wang, W., Xu, Z., Zhang, K.: Pf-lrm: Pose-free large reconstruction model for joint pose and shape prediction. arXiv preprint arXiv:2311.12024  (2023)

\bibitem{wang2020learning}
Wang, Q., Zhou, X., Hariharan, B., Snavely, N.: Learning feature descriptors using camera pose supervision. In: Proc. European Conference on Computer Vision (ECCV) (2020)

\bibitem{wang2023dust3r}
Wang, S., Leroy, V., Cabon, Y., Chidlovskii, B., Revaud, J.: Dust3r: Geometric 3d vision made easy. arXiv preprint arXiv:2312.14132  (2023)

\bibitem{wang2024prolificdreamer}
Wang, Z., Lu, C., Wang, Y., Bao, F., Li, C., Su, H., Zhu, J.: Prolificdreamer: High-fidelity and diverse text-to-3d generation with variational score distillation. Advances in Neural Information Processing Systems  \textbf{36} (2024)

\bibitem{wang2004ssim}
Wang, Z., Bovik, A.C., Sheikh, H.R., Simoncelli, E.P.: Image quality assessment: from error visibility to structural similarity. IEEE transactions on image processing  \textbf{13}(4),  600--612 (2004)

\bibitem{watson2022novel}
Watson, D., Chan, W., Martin-Brualla, R., Ho, J., Tagliasacchi, A., Norouzi, M.: Novel view synthesis with diffusion models. arXiv preprint arXiv:2210.04628  (2022)

\bibitem{weyand2020GLDv2}
Weyand, T., Araujo, A., Cao, B., Sim, J.: {Google Landmarks Dataset v2 - A Large-Scale Benchmark for Instance-Level Recognition and Retrieval}. In: Proc. CVPR (2020)

\bibitem{wiles2020synsin}
Wiles, O., Gkioxari, G., Szeliski, R., Johnson, J.: Synsin: End-to-end view synthesis from a single image. In: CVPR. pp. 7467--7477 (2020)

\bibitem{wu2023reconfusion}
Wu, R., Mildenhall, B., Henzler, P., Park, K., Gao, R., Watson, D., Srinivasan, P.P., Verbin, D., Barron, J.T., Poole, B., et~al.: Reconfusion: 3d reconstruction with diffusion priors. arXiv preprint arXiv:2312.02981  (2023)

\bibitem{Wu2021Towers}
Wu, X., Averbuch-Elor, H., Sun, J., Snavely, N.: {Towers of Babel}: {C}ombining images, language, and {3D} geometry for learning multimodal vision. In: ICCV (2021)

\bibitem{depthanything}
Yang, L., Kang, B., Huang, Z., Xu, X., Feng, J., Zhao, H.: Depth anything: Unleashing the power of large-scale unlabeled data. In: CVPR (2024)

\bibitem{yao2020blendedmvs}
Yao, Y., Luo, Z., Li, S., Zhang, J., Ren, Y., Zhou, L., Fang, T., Quan, L.: Blendedmvs: A large-scale dataset for generalized multi-view stereo networks. Computer Vision and Pattern Recognition (CVPR)  (2020)

\bibitem{yeshwanth2023scannet++}
Yeshwanth, C., Liu, Y.C., Nie{\ss}ner, M., Dai, A.: Scannet++: A high-fidelity dataset of 3d indoor scenes. In: Proceedings of the IEEE/CVF International Conference on Computer Vision. pp. 12--22 (2023)

\bibitem{yu2021pixelnerf}
Yu, A., Ye, V., Tancik, M., Kanazawa, A.: pixelnerf: Neural radiance fields from one or few images. In: Proceedings of the IEEE/CVF Conference on Computer Vision and Pattern Recognition. pp. 4578--4587 (2021)

\bibitem{yu2023wonderjourney}
Yu, H.X., Duan, H., Hur, J., Sargent, K., Rubinstein, M., Freeman, W.T., Cole, F., Sun, D., Snavely, N., Wu, J., et~al.: Wonderjourney: Going from anywhere to everywhere. arXiv preprint arXiv:2312.03884  (2023)

\bibitem{yu2023mvimgnet}
Yu, X., Xu, M., Zhang, Y., Liu, H., Ye, C., Wu, Y., Yan, Z., Liang, T., Chen, G., Cui, S., Han, X.: Mvimgnet: A large-scale dataset of multi-view images. In: CVPR (2023)

\bibitem{zhang2018perceptual}
Zhang, R., Isola, P., Efros, A.A., Shechtman, E., Wang, O.: The unreasonable effectiveness of deep features as a perceptual metric. In: CVPR (2018)

\bibitem{RealEstate10k}
Zhou, T., Tucker, R., Flynn, J., Fyffe, G., Snavely, N.: Stereo magnification: Learning view synthesis using multiplane images. ACM Trans. Graph. (Proc. SIGGRAPH)  \textbf{37} (2018), \url{https://arxiv.org/abs/1805.09817}

\bibitem{zhou2016view}
Zhou, T., Tulsiani, S., Sun, W., Malik, J., Efros, A.A.: View synthesis by appearance flow. In: European Conference on Computer Vision (2016)

\bibitem{zhou2023sparsefusion}
Zhou, Z., Tulsiani, S.: Sparsefusion: Distilling view-conditioned diffusion for 3d reconstruction. In: Proceedings of the IEEE/CVF Conference on Computer Vision and Pattern Recognition. pp. 12588--12597 (2023)

\end{thebibliography}

\appendix
\section*{Appendix}
\section{Visualizations of Dataset Characteristics}
\begin{figure}[h]
    \centering
    \includegraphics[width=0.99\textwidth,trim={0 0 0 0},clip]{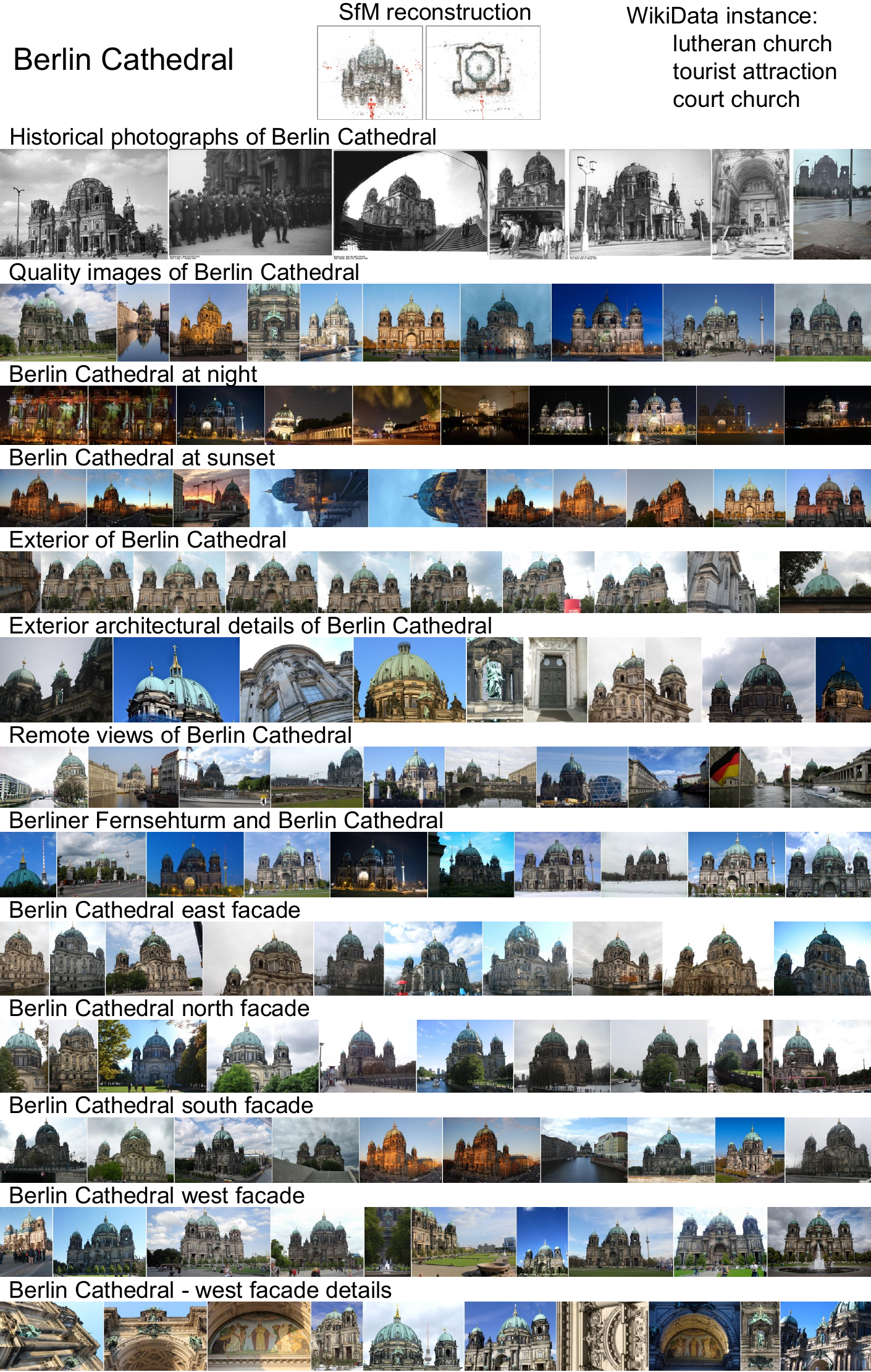}
    \caption{
    Registered images of Berlin Cathedral, organized by Wikimedia Commons subcategories. 
    Each text label corresponds to a subcategory (possibly nested) of the main category.
    }
    \label{fig:appx-scene}
\end{figure}

\begin{figure}[h]
    \centering
    \includegraphics[width=0.99\textwidth,trim={0 0 0 0},clip]{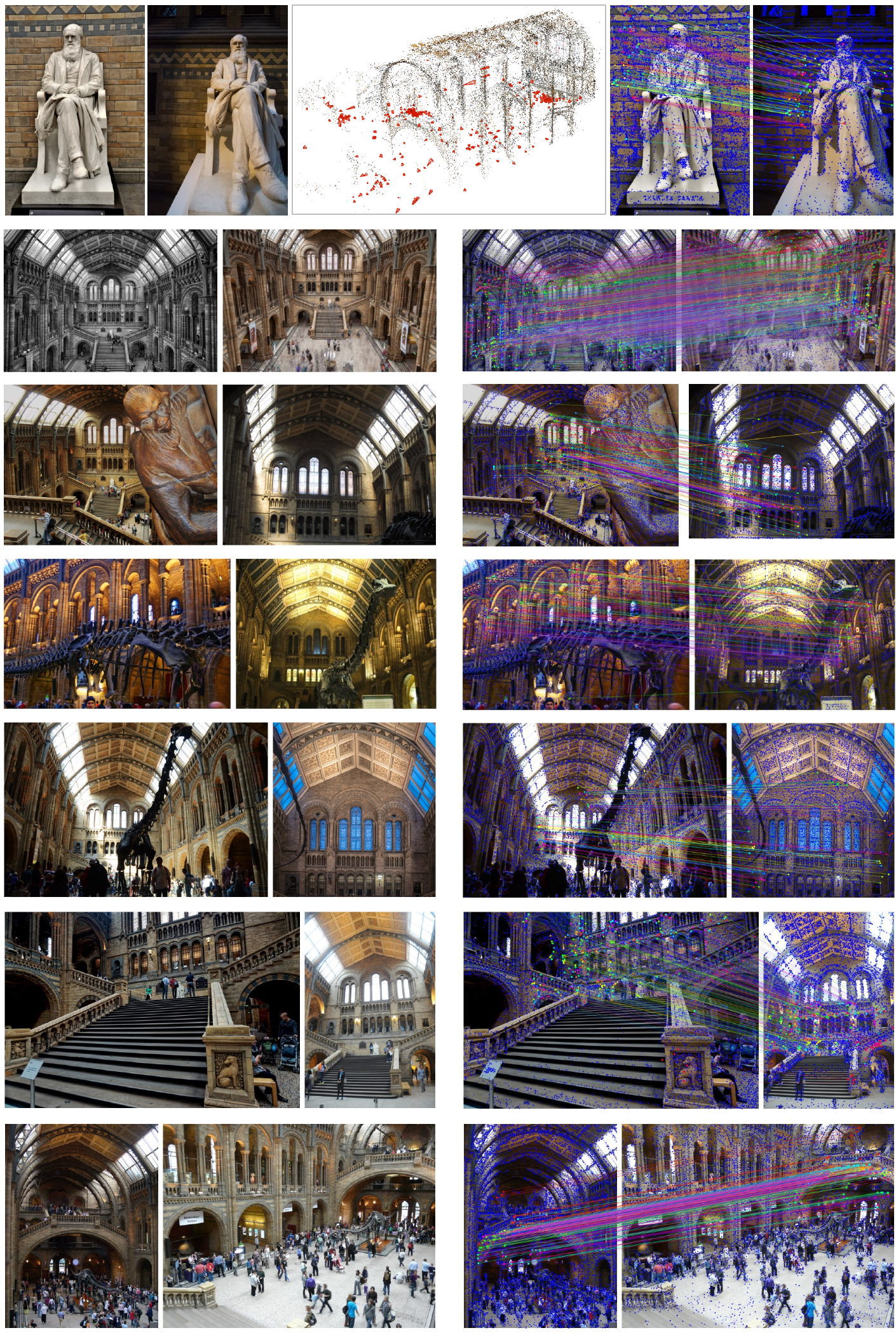}
    \caption{
    Feature matching visualization of the Natural History Museum's interior in London, with the SfM reconstruction shown in the middle of the top row. 
    On the left, selected image pairs are shown, and on the right, the extracted keypoints (in blue) and their matches (in color-coded) from the two-view geometry table in COLMAP~\cite{schoenberger2016sfm} database are displayed.
    }
    \label{fig:appx-twoview}
\end{figure}

We provide additional figures to convey the wealth of information in the MegaScenes dataset. In \cref{fig:appx-scene}, we highlight the diversity of images registered to a reconstruction, Wikidata class information, and image subcategories for an outdoor scene. In \cref{fig:appx-twoview}, we show examples of image pairs with calculated two-view geometries for an indoor scene.

\section{Details for Dataset Curation}

\subsection{Processing Wikidata Entries for Scene Identification}
\label{sec:appx-process_Wikidata}
In the first dataset curation step, ``Identifying Scenes,'' we collect raw Wikidata entries from broad categories that link to Wikimedia Commons categories. Each Wikidata entry points to some Wikimedia Commons category; we take the set of these Wikimedia Commons categories to use as scenes. Before we download images from these categories, we do additional cleaning steps to determine which Wikimedia Commons categories to use.

\medskip
\noindent \textbf{Filtering Wikidata Entries based on Cyclic Links.}
Some collected Wikidata entries point to broad Wikimedia Commons categories, 
like \href{https://commons.wikimedia.org/wiki/Category:Fountains}{Fountains} 
or \href{https://commons.wikimedia.org/wiki/Category:Cultural_heritage_monuments_in_Toropetsky_District}{Cultural heritage monuments in Toropetsky District},
unsuitable to use as single scenes. We note that a Wikidata entry points to some Wikimedia Commons category, and a Wikimedia Commons category points towards some, but not necessarily the same, Wikidata entry. To clean the set of Wikimedia Commons categories as described above, we ensure that there is a cyclic link between the Wikimedia Commons Category and its corresponding Wikidata entry. A Wikimedia Commons category like \href{https://commons.wikimedia.org/wiki/Category:Fountains}{Fountains} will point to a Wikidata entry about fountains, and not the original Wikidata entry that pointed to the \href{https://commons.wikimedia.org/wiki/Category:Fountains}{Fountains} Wikimedia Commons category.

\medskip
\noindent \textbf{Filtering Wikidata Entries based on GLAM Instances.} We find that some categories related to galleries, libraries, archives, and museums (GLAM) contain many images that are unhelpful in 3D reconstructions, such as 2D scans of paintings or text. To minimize the number of such 2D scans, we ignore all Wikidata entries that are \textit{exclusively} GLAM instances. We keep the Wikidata entries that are also instances of at least one other unrelated class, as these are more likely to have images that are not exclusively of 2D scans.

\subsection{Subcategory Recursion when Downloading Images}
In the second dataset curation step, ``Downloading Images from Scenes,'' we download images from every Wikimedia Commons category deemed a scene. From the original Commons category, we recurse a maximum depth of four subcategories and download all associated images.

However, some subcategories are unrelated to the original scene, and we want to avoid downloading images unhelpful in 3D reconstruction. For instance, if a subcategory begins with ``People associated with\ldots'', then the subcategory will link to Wikimedia Commons pages that contain images of individuals, rather than the original scene.

To fix the above issue, we define two conditions that must be met in order to recurse into a related subcategory. First, we create a list of excluded keywords that the subcategory must not contain. We curate this list by experimentally finding common diverging subcategories, which includes keywords like ``People associated with\ldots''.
Second, the subcategory must contain a substring that includes one of the following names:
\begin{itemize}
    \item[$\bullet$] The name of the original Wikimedia Commons category
    \item[$\bullet$] The name of the Wikidata entity associated with the original Wikimedia Commons category
    \item[$\bullet$] Any alias recorded on the Wikidata entity associated with the original Wikimedia Commons category
\end{itemize}

\subsection{Details on Reconstruction and Cleaning}
We elaborate on the third curation step, ``Reconstructing Scenes with SfM and Cleaning Reconstructions.''

\medskip
\noindent \textbf{Reconstruction Orientation Alignment.} The reconstructions created by COLMAP~\cite{schoenberger2016sfm} are not necessarily aligned to the real world. For instance, the gravity axis as seen in input images may not align with the down-axis of the reconstruction's coordinate system. Thus, we orient all reconstructions using COLMAP's implementation of Manhattan world alignment. This process aligns the sparse reconstruction using the Manhattan World assunption~\cite{manhattanworld}, which assumes that most surfaces are aligned along the three major axes. 

\medskip
\noindent \textbf{Cleaning Watermarked Reconstructions.} Some scenes have many images with watermarks. COLMAP finds spurious matches between the watermarks of two images, which result in incorrect sparse reconstructions. We assume that most images uploaded to Wikimedia Commons have watermarks that are non-destructive and are near the borders of the image. To fix this issue, we mask all keypoints within a certain distance near the image border before we rerun COLMAP's feature matching and reconstruction phases. Specifically, we target all scenes where at least 10 percent of the inlier pairs are ``watermark pairs'' as labeled in COLMAP's output database. We find that a border defined by 5 percent of the image diagonal is able to mask watermarks in most images.

\medskip
\noindent \textbf{Using Doppelgangers to Clean Reconstructions.} As discussed in the main paper, we use the Doppelgangers~\cite{cai2023doppelgangers} pipeline to fix incorrect SfM reconstructions caused by visual ambiguities. Incorrect reconstructions arise from false correspondences between image pairs, such as in photos that depict different surfaces that are similar in appearance. Doppelgangers uses a binary classifier that predicts the likelihood of whether an input image pair should be matched; the input pairs to SfM are filtered by passing them through the Doppelgangers classifier. After we filter the image pairs, we rerun COLMAP's reconstruction phase. We find that the default threshold of keeping pairs with a confidence score of $\ge0.8$ is able to correctly disambiguate most scenes. If thresholding at 0.8 is unsuccessful, we try increasing thresholds until the reconstruction is correct. 

\section{Additional Details on Dataset Statistics}

\subsection{Wikidata Classes to Identify Scenes}
We show the Wikidata classes we use to identify scenes in our dataset curation process in Table \ref{tab:appx-wikidata_class}. Refer to \cref{sec:appx-process_Wikidata} for how the 660K Wikidata entries are filtered into 430K scenes.
\begin{table}
    \centering
    \begin{tabular}{lr}
    \toprule
    Wikidata Class                   & Entry Count  \\
    \midrule
    religious building               & 233,253       \\
    monument                         & 75,196        \\
    tourist attraction               & 55,051        \\
    museum                           & 40,653        \\
    landmark                         & 34,950        \\
    bridge                           & 33,207        \\
    chapel                           & 29,866        \\
    commercial building              & 24,859        \\
    public building                  & 24,227        \\
    shrine                           & 22,055        \\
    tower                            & 17,915        \\
    square                           & 13,817        \\
    statue                           & 10,872        \\
    palace                           & 10,237        \\
    Catholic church building         & 8,792         \\
    fountain                         & 5,496         \\
    high-rise building               & 4,083         \\
    Eastern Orthodox church building & 3,782         \\
    cathedral                        & 3,326         \\
    mosque                           & 3,093         \\
    house of prayer                  & 3,092         \\
    library building                 & 742          \\
    arch                             & 349          \\
    gurdwara                         & 57           \\
    \midrule
    \textbf{Total}                   & \textbf{659,024}
    \\ \bottomrule
    \end{tabular}

    \vspace{7pt}
    \caption{
    Wikidata classes that have been selected to identify a set of scenes. Some Wikidata entries may be present in multiple classes. Multiple Wikidata entries may link to the same Wikimedia Commons category.
    }
    
    \label{tab:appx-wikidata_class}
\end{table}

\subsection{Scene Overlap with Google Landmarks Dataset V2}
While MegaScenes and Google Landmarks Dataset V2 (GLDv2)~\cite{weyand2020GLDv2} both source images from Wikimedia Commons, we find that neither dataset has a majority overlap with the other in terms of categories used as scenes. MegaScenes contains 430K categories as scenes, while GLDv2 contains 213K categories as scenes; there are 74K scenes that are found in both datasets. This means that MegaScenes has 356K scenes not found in GLDv2, and GLDv2 has 139K scenes not found in MegaScenes. We attribute this to differing data curation methods for both datasets: GLDv2 queries the Google Knowledge Graph, while MegaScenes utilizes Wikidata.

\section{Details for Novel View Synthesis Experiments}

\subsection{Data Setup for Evaluation}
\medskip
\noindent \textbf{DTU.} We use the test split of 15 scenes from previous work~\cite{sargent2023zeronvs} on the DTU dataset~\cite{dtu}. Each scene contains 49 images with the same exact array of camera positions, and we pick two reference locations that are across from each other. See the supplement for more details. For each reference image, we exhaustively form pairs with all other images in the scene. This results in 95$\times$2 pairs (we count $(a,b)$ and $(b,a)$ as separate pairs) per scene, for a total of 2,850 pairs.

\medskip
\noindent \textbf{Mip-NeRF 360.} We use all 9 scenes from the original Mip-NeRF 360 dataset~\cite{barron2022mipnerf360}. We leverage how the images in these scenes form a 360 degree orbit about a central location to identify reference images. We align the provided COLMAP sparse point cloud using the Manhattan world assumption~\cite{manhattanworld}, then sort the images by increasing viewing direction angle on the XZ plane. We sample ten evenly distributed images from this sorted list. For each reference image, we pair it with all images that share at least 50 3D points in the sparse point cloud. This results in a total of 15,862 pairs across all scenes.

\medskip
\noindent \textbf{RealEstate10K (Re10K).} We adopt the pair sampling strategy from Synsin~\cite{wiles2020synsin} for Re10K~\cite{RealEstate10k}, which selects a reference and target video frame no more than 30 frames apart. To identify more challenging frames, the authors choose pairs with an angular change greater than 5$^\circ$ and a positional change greater than 0.15, whenever possible. 
We take the intersection of these pairs and ZeroNVS'~\cite{sargent2023zeronvs} held out set (since ZeroNVS was trained on RE10K). In total, we obtain 644 pairs across 163 clips.

\subsection{Finetuning and Inference}
For all finetuning, we use 6 NVIDIA A6000 GPUs with a total batch size of 1656 until the metrics of the validation set stop improving. We finetune ZeroNVS~\cite{sargent2023zeronvs} for 30,000 iterations which takes 1-2 days. We finetune Zero-1-to-3~\cite{liu2023zero} for 75,000 iterations. 
During inference, we use 50 DDIM~\cite{DDIM} steps for all qualitative and quantitative results. We use classifier-free guidance scales~\cite{ho2022classifier} of 3.0 for all finetuned models and \textit{SD-inpainting}. We observed less realistic generations with little to no improvements in metrics when setting a higher scale. For \textit{Zero-1-to-3 (released)} and \textit{ZeroNVS (released)}, we use a scale of 7.5 following the default in ZeroNVS, as the models match the target poses better at a higher scale. 

Our method (denoted \textit{Ours} in the main paper) is finetuned from ZeroNVS and combines the warped images and extrinsic matrices as conditions. Similar to Zero-1-to-3 and ZeroNVS, we pass the target and reference images through a pretrained VAE to obtain their latents, each with shape (4, 32, 32). We downsample the warped image from (3, 256, 256) to (3, 32, 32) and concatenate it with the target and reference latents, so that our input to the first layer of the diffusion model has shape (11, 32, 32). The conditioning of the extrinsic matrices is exactly follows ZeroNVS. We scale the translation vector of the extrinsic matrix from COLMAP by the 20th quantile of the aligned depth (Fig 4 of the main paper).

For \textit{SD-inpainting}, we use the checkpoint \textit{sd-v1-5-inpaint.ckpt} (\url{https://huggingface.co/runwayml/stable-diffusion-inpainting}) from Runway ML. Since the model is trained on 512x512 images, we use 512x512 images and 64x64 latents to match. Then, we downsample the outputs to 256x256 for qualitative results and calculating metrics. 

\section{Additional Qualitative Results and Comparisons}

We show additional qualitative results uniformly sampled from our test set in the PDF file named \textit{qualitative results.pdf}. 
We will release data, seeds, and models for reproduction. In virtually all cases, finetuned models \textit{(MS)} outperform base models in terms of pose consistency and realism. \textit{Ours} generally follows the target pose more closely than \textit{ZeroNVS (MS)} and \textit{Zero-1-to-3 (MS)}. 

We also finetune ZeroNVS on MegaDepth~\cite{MegaDepthLi18}, denoted \textit{ZeroNVS (MD)}. MegaDepth is the most similar dataset to MegaScenes, consisting of diverse internet photos with COLMAP reconstructions. However, MegaDepth is of a significantly smaller scale. For the task of novel view synthesis, we process MegaDepth through the same process as described in Section 4.1 of the main paper, and obtain a total of 368,028 training pairs and 181 scenes, roughly 6 times fewer pairs, and 180 times fewer scenes. We find that \textit{ZeroNVS (MS)} generally outperforms \textit{ZeroNVS (MD)} in terms of both following the target pose and realism, especially of less common scenes. 

\section{Videos}
We show videos on our project page. Videos contain sequential frames and make it easy to visualize the pose consistencies from frame to frame. We also include an example of autoregressive generation, where we take the last view of a generated sequence as the first frame of a new sequence. However, since each image is sampled independently, long sequences eventually drift. In the future, adding temporal constraints or directly generating multiple frames could solve this issue. 

\section{Broader Impact}
This work primarily focuses on the task of novel view synthesis, and similar to other generative models, present risks such as the potential for generating misleading or harmful content. It is essential to develop robust frameworks for ethical use.

\end{document}